\pdfoutput=1

\documentclass[11pt]{article}

\usepackage[]{acl}

\usepackage{times}
\usepackage{latexsym}
\usepackage{graphicx}
\usepackage{multirow}
\usepackage{tabularx}
\usepackage{booktabs}
\usepackage{amsmath}
\usepackage{amssymb}
\usepackage{lipsum}
\newcommand{\modelname}{\textsc{Pairwise\-RL}}

\usepackage[T1]{fontenc}

\usepackage[utf8]{inputenc}

\usepackage{microtype}

\title{
\vspace*{-0.5in}
{{\small \hfill *SEM'22}\\
\vspace*{.25in}} 
Pairwise Representation Learning for Event Coreference 
}

\author{Xiaodong Yu$^{1}$ \enspace\enspace\enspace Wenpeng Yin$^{2}$ \enspace\enspace\enspace Dan Roth$^{1}$  \\
$^{1}$University of Pennsylvania \enspace\enspace\enspace $^{2}$Temple University\\
  {\tt \small \{xdyu, danroth\}@seas.upenn.edu  \enspace\enspace\enspace wenpeng.yin@temple.edu} 
}

\begin{document}
\maketitle
\begin{abstract}

Natural Language Processing tasks such as resolving the coreference of events require understanding the relations between two text snippets.
These tasks are typically formulated as (binary) classification problems over independently induced representations of the text snippets.
In this work, we develop a Pairwise Representation Learning (\modelname) scheme for the event mention pairs, in which we jointly encode a pair of text snippets so that the representation of each mention in the pair is induced in the context of the other one. Furthermore, our representation supports a finer, structured representation of the text snippet to facilitate encoding events and their arguments. 
We show that \modelname, despite its simplicity, outperforms the prior state-of-the-art event coreference systems on both cross-document and within-document event coreference benchmarks. We also conduct in-depth analysis in terms of the improvement and the limitation of pairwise representation so as to provide insights for future work. \footnote{Our code is available at \url{http://cogcomp.org/page/publication_view/979}}

\end{abstract}

\section{Introduction}

In this work, we study the event coreference resolution problem.
Event coreference resolution is commonly modeled as a binary classification problem over independently induced representations on the text snippets of each event mention  \cite{lee2012joint, barhom-etal-2019-revisiting}.\footnote{Some work maps the two mentions into a single matching score, e.g., \cite{barhom-etal-2019-revisiting}; this can be treated as a special case of binary classification.}
Understanding the relations between two text snippets is the essential part in the tasks. In this work, we argue that the representations of prior work are not expressive enough to learn the pairwise relations due to the following two reasons: \\ (i) \textit{Counterpart Unawareness}. 
The relationship between two mentions can be different in different contexts. To address different scenarios, it is better for each mention to ensure that its representation is aware of what its counterpart's representation.  
%
However, most early work induces mention representations independently by extracting features only from the 
sentence that contains 
the mention, without using the context of the other mention \cite{barhom-etal-2019-revisiting, huang2019improving}. Some more recent work 
attempts to encode the whole document to represent each mention \cite{lee-etal-2017-end, catton2020stream}. This is beneficial 
for short documents, since the representation of each mention will also include information from the context of the other candidate mention. However, this is not sufficient for cross-document settings, when the comparison is, for example, between two event mentions that appear in separate documents. In this case even encoding large pieces of text leave the candidate mention representations independent of each other.
\\
(ii) \textit{Unstructured representation learning.} An event mention consists of multiple arguments that describe the event: who, when, where, etc. When determining the relationship of two event mentions, the mismatch of some arguments could be decisive. Consider the following two sentences $s_1$ and $s_2$ (event trigger is \underline{\textbf{underlined}}; argument \#0 is in \textcolor{blue}{blue}, location is in \textcolor{purple}{purple})\\

\fbox{%
  \begin{minipage}{0.92\linewidth}
    \small $s_1$: ``Over \textcolor{blue}{69,000 people} \underline{\textbf{lost}} their lives
    in the quake, including 68,636 in \textcolor{purple}{Sichuan}.''\\
    $s_2$: ``Up to \textcolor{blue}{6,434 people} \underline{\textbf{lost}} their lives 
    in Kobe earthquake and about 4,600 of them were from \textcolor{purple}{Kobe}.''
  \end{minipage}
}
\\



These two events ``lost'' are not the same events because the earthquake in Sichuan and the earthquake in Kobe are two different earthquakes, and Sichuan and Kobe do not have any geographic overlap. The mismatch of the locations ``Sichuan'' and ``Kobe'' may be enough to determine that the two events are different from each other without even considering the rest of the sentence.
Most prior work encodes all of the arguments into a single distributed representation vector and just compares the overall vector representations of two mention triggers.
Although contextual representation could encode all of the arguments' information, this is less optimal than explicitly representing all of the arguments, thus making 
it easier for the model to conduct fine-grained reasoning over each of the argument.

To address the drawbacks of prior representations, we propose 
\textit{pairwise representation learning}
(\modelname). \modelname\enspace alleviates the aforementioned two limitations with two designs:

\textbf{Pairwise representation learning.} 
We suggest treating a mention pair, rather than a single mention, as the object for the representation learning. We encode the two mentions' sentences as a whole sequence so that one sentence's token representation is able to interact with the other sentence's from the very beginning.
This is advantageous over learning two separate and independent representations because it allows for learning how compatible one mention is with the other mention's context.

\textbf{Structured representation learning.} 
The observation that mismatching arguments are critical to making the coreference decision indicates that using a single combined representation for all of the arguments could be less informative for cross-mention comparison.
In this work, we explicitly represent all the arguments, and compare each argument separately.

To our knowledge, this is the first work that applies pairwise representation learning to event coreference problems. We report our performance on both within-document and cross-document event coreference benchmarks.
We show that \modelname, despite its simplicity, clearly surpasses more complex state-of-the-art event coreference systems on two most popular benchmarks ECB+ \cite{cybulska2014using} and KBP17 \cite{Getman2015OverviewOL}. We also conduct in-depth analysis in terms of the improvement and the limitation of pairwise representation so as to provide insights for future work.



\section{Related Work}
In this section, we  discuss  prior representation learning approaches for event coreference and how pairwise representation learning has been used in other NLP problems.
\paragraph{Event Coreference.}

Earlier work uses hand-engineered event features to represent events \cite{chen2009event, bejan-harabagiu-2010-unsupervised}. 

Most recent neural models use contextual embedding and character-based embedding of event triggers with some pairwise features to represent events \cite{kenyon-dean-etal-2018-resolving, huang2019improving, catton2020stream}. These works do not use argument information, and expect the contextual embedding includes all the necessary information. 

Argument information has been integrated into event representations either by encoding some string-level features \cite{peng-etal-2016-event,choubey-huang-2017-event} or by entity-level coreference co-training \cite{lee2012joint,barhom-etal-2019-revisiting}.

In contrast, our representation learning of events has a unified system to encode the event triggers and the argument entities, which avoids the costly co-training while learning more advanced features that express the arguments on their own and their interactions with the event triggers.

\paragraph{Pairwise Representation Learning in Other NLP Tasks.}
Pairwise representation learning has been widely adopted to model the relationships of two pieces of text. The main goal is to learn contextualized sentence representations. Earlier systems commonly implement with attention mechanisms in recurrent \cite{DBLPnKGEKSB15}, convolutional \cite{DBLPnS18} or Transformer-style \cite{DBLPwaniSPUJGKP17} neural networks  to deal with text
generation, such as neural machine translation \cite{DBLPhdanauCB14}, 
document reconstruction \cite{DBLPLiLJ15}, and
document summarization  \cite{DBLPtiZSGX16};
machine comprehension \cite{DBLPnKGEKSB15}, textual entailment \cite{DBLPtaschelGHKB15,Devlin2019BERTPO}, etc.


In this work, we develop the pairwise representation learning for modeling the relationship of two mentions within two separate sentences rather than the relationship of the two sentences themselves. To the best of our knowledge, we are the first to (i) study pairwise representation for event pairs by letting two mentions learn from each other's context from the beginning 
\footnote{\cite{DBLPZengJGGC20} uses a similar method, and is a contemporary work with ours.}
, and (ii) build structured representation between events by fine-grained argument reasoning, without any hand-engineered features.

\section{\modelname~for Coreference}
\label{sec:model}
\begin{figure}[t]
\centering
\includegraphics[width=7.5cm]{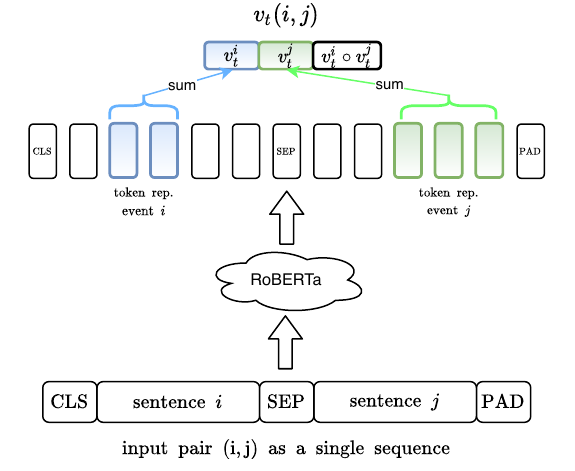}
\caption{\modelname\enspace learns the trigger-only pair-wise representation. $v_t^i$ (resp. $v_t^j$) is the contextualized representation vector for the trigger in event $i$ (resp. $j$). The whole trigger-based event pair ($i$, $j$) is denoted by $v_t(i,j)$ which is the concatenation: [$v_t^i$, $v_t^j$, $v_t^i\circ v_t^j$].}
\label{fig:model}
\end{figure}

 
\modelname~takes two sentences containing each mention as the input and outputs a score indicating how likely the two mentions refer to the same event. 
Given the mention pair $e_i$ and $e_j$ with their arguments [arg0; arg1; loc; time], as shown in Fig \ref{fig:model}, we concatenate the sentences of $e_i$ and $e_j$, and encode the concatenated sentence using RoBERTa \cite{liu2019roberta}. 
After encoding each token of the sequence to a representation vector, we sum up the token representations of the mention span as the representations for event trigger and event arguments respectively: $v_{\mathrm{t}}$ for event trigger, $v_{\mathrm{arg0}}$/$v_{\mathrm{arg1}}$ for argument \#0 or \#1, $v_{\mathrm{loc}}$ for location and $v_{\mathrm{time}}$ for time. 

\begin{figure}[t]
\centering
\includegraphics[width=7.5cm]{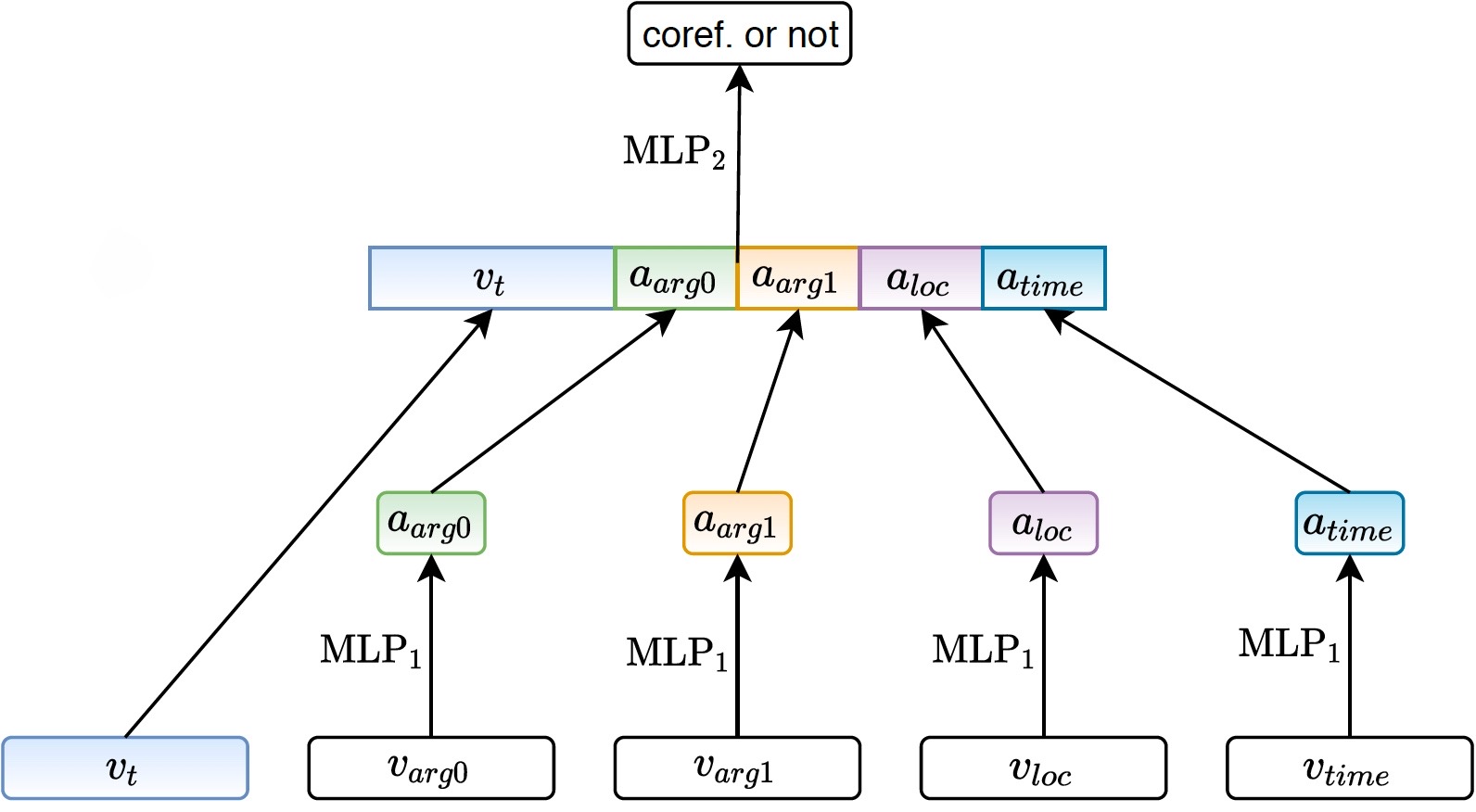}
\caption{The full reasoning process in \modelname. The final \modelname\enspace representation is the concatenation of the trigger's representation and four feature values, each coming from a mention argument. }
\label{fig:pairstruc}
\end{figure}

Next, we conduct fine-grained coreference reasoning, as Figure \ref{fig:pairstruc} shows. The goal is to let each role of event arguments learn its contribution to the final task. 
For each role, where $\mathrm{role} \in $ \{t, arg0, arg1, loc, time\}, we first build the following role-wise representation:
\begin{equation}
    v_{\mathrm{role}}(i,j)= [v^i_{\mathrm{role}}, v^j_{\mathrm{role}}, v^i_{\mathrm{role}} \circ v^j_{\mathrm{role}}]
\end{equation}
where $\circ$ is element-wise multiplication. Because these four arguments may not always exist in the local context, if one of the role is missing, then the corresponding $v^i_{\mathrm{role}}$ will be a zero vector.

We keep the $v_{\mathrm{t}}$ as the main representation in \modelname, and let each of the remaining four arguments contribute a feature value indicating their own decisiveness. The feature value is learnt with a  multi-layer perceptron (MLP) as follows:
\begin{equation}
    a_{\mathrm{role}}(i,j) = \mathrm{MLP}_1(v_{\mathrm{role}}(i,j))
\end{equation}
where ``role'' refers to mention arguments other than the trigger, $\mathrm{MLP}_1$ has four layers and the output of $\mathrm{MLP}_1$ is a single scalar as the argument feature value.
As a result, the final representation \modelname~for event coreference is:
\begin{equation}\label{eq:finalrep}
 v(i,j) = [v_t(i,j), a_{\mathrm{arg0}}, a_{\mathrm{arg1}}, a_{\mathrm{loc}}, a_{\mathrm{time}}]
\end{equation}
Since entities do not have arguments, the final representation \modelname~for entity coreference is:
\begin{equation}\label{eq:finalrepentity}
 v(i,j) = v_t(i,j)
\end{equation}
Once obtaining the pairwise representation $v(i,j)$, another four-layer MLP, as shown in Figure \ref{fig:pairstruc}, will act as a binary classifier (i.e., is coreferential or not)
\begin{equation}\label{eq:MLP2}
p(i,j) =\mathrm{Softmax}(\mathrm{MLP_2}(v(i,j)))
\end{equation}
where $p(i,j)[0]$ is the probability that the two mentions $i$ and $j$ are coreferential.

\begin{table}[t]
\centering
\begin{tabular}{l|rrr}
\toprule
 & Train & Dev & Test \\
\midrule
Topics & 25 & 8 & 10 \\
Documents & 574 & 196 & 206 \\
Sentences & 1,037 & 346 & 457 \\
Event mentions & 3,808 & 1,245 & 1,780 \\
Event Singletons & 1,116 & 280 & 623 \\
Event Clusters & 1,527 & 409 & 805 \\
Entity mentions & 4,758 & 1,476 & 2055 \\
Entity Singletons & 472 & 125 & 196 \\
Entity Clusters & 1,286 & 330 & 608 \\
\bottomrule
\end{tabular}
\caption{ECB+ statistics. We follow the data split by \citet{cybulska2015bag}: \textit{train}: 1, 3, 4, 6-11, 13-17, 19-20, 22, 24-33; \textit{dev}: 2, 5, 12, 18, 21, 23, 34, 35; \textit{test}: 36-45. Event/Entity Clusters include singletons.} 
\label{tab:ecbstat}
\end{table}

\begin{table*}[ht]
\setlength{\tabcolsep}{5pt}
\centering
\begin{tabular}{lccccccccccccccc}
\toprule
\multirow{2}{*}{ } & & \multicolumn{3}{c}{MUC} & & \multicolumn{3}{c}{ $\mathrm{B}^3$} & & \multicolumn{3}{c}{$\mathrm{CEAF}_e$} & & CoNLL \\ 
\cline{3-5}
\cline{7-9}
\cline{11-13}
\cline{15-15}
Model & & R & P & F1 & & R & P & F1 & & R & P & F1 & & F1 \\
\midrule
same head lemma && 76.5 & 79.9 & 78.1 && 71.7 & 85 & 77.8 && 75.5 & 71.7 & 73.6 && 76.5 \\
\citet{barhom-etal-2019-revisiting} && 77.6 & 84.5 & 80.9 && 76.1&85.1 &80.3&& 81&73.8&77.3&&79.5  \\
\citet{catton2020stream} && 85.1 & 81.9 & 83.5 && 82.1 & 82.7 & 82.4 && 75.2 & 78.9 & 77.0 && 81.0  \\
\midrule

Unpaired & \\
\enspace\enspace\enspace\enspace Unstructured && 81.7 & 84.4 & 83.1 && 79.8 & 86.3 & 82.9 && 79.6 & 76.7 & 78.1 && 81.3\\
\enspace\enspace\enspace\enspace Structured && 84.6 & 84.6 & 84.6 && 83.6 & 84.2 & 83.9 && 80.2 & 80.2 &  80.2 && 82.9 \\

Pairwise  && \\
\enspace\enspace\enspace\enspace Unstructured && 91.6 & 83.1 & \textbf{87.2} && 89.4 & 81.1 & 85.1 && 75.0 & 85.5 & 79.9 && 84.0 \\
\enspace\enspace\enspace\enspace Structured && 88.1 & 85.1 & 86.6 && 86.1 & 84.7 & \textbf{85.4} && 79.6 & 83.1 & \textbf{81.3} && \textbf{84.4} \\
\enspace\enspace\enspace\enspace $\textrm{Structured}_{\mathrm{BERT}}$ && 87.4 & 81.4 & 84.3 && 85.7 & 80.2 & 82.9 && 73.7 & 80.9 & 77.1 && 81.4 \\
\bottomrule
\end{tabular}
\caption{Cross-document event coreference performance on ECB+. All the models use gold mentions and predicted topics. ``Unstructured'' means the model only uses the representation of the event trigger. ``Structured'' means the model uses the structured representation of arguments. ``Unpaired'' is the baseline model without pairwise representation. ``Pairwise'' is the model using pairwise representation. $\mathrm{Structured}_{\mathrm{BERT}}$ means this baseline model uses BERT \cite{Devlin2019BERTPO} as contextual embeddings instead of RoBERTa. Details in Sec \ref{sec:crossdocevent}.} 
\label{tab:ecbresult}
\end{table*}

\begin{table*}[h!]
\setlength{\tabcolsep}{5pt}
\centering
\begin{tabular}{lccccccccccccccc}
\toprule
\multirow{2}{*}{ } & & \multicolumn{3}{c}{MUC} & & \multicolumn{3}{c}{ $\mathrm{B}^3$} & & \multicolumn{3}{c}{$\mathrm{CEAF}_e$} & & CoNLL \\ 
\cline{3-5}
\cline{7-9}
\cline{11-13}
\cline{15-15}
Model & & R & P & F1 & & R & P & F1 & & R & P & F1 & & F1 \\
\midrule
\citet{barhom-etal-2019-revisiting} && 78.6 & 80.9 & 79.7 && 65.5&76.4 &70.5&& 65.4&61.3&63.3&&71.2  \\
\citet{catton2020stream} && 85.7 & 81.7 & 83.6 && 70.7 & 74.8 & 72.7 && 59.3 & 67.4 & 63.1 && 73.1  \\
\midrule
\modelname  && 92.3 & 86.8 & \textbf{89.5} && 82.1 & 81.0 & \textbf{81.5} && 68.0 & 80.2 & \textbf{73.6} && \textbf{81.5} \\
\bottomrule
\end{tabular}
\caption{Cross-document Entity coreference performance on ECB+. All the models evaluate on gold mentions and predicted topics.} 
\label{tab:ecbentityresult}
\end{table*}

\section{Experiments}
We apply \modelname~to cross-document  and within-document event coreference problems. 

\subsection{Cross-document Event Coreference}
\label{sec:crossdocevent}

\paragraph{Dataset} We use the ECB+ \cite{cybulska2014using} corpus to train and test our model. ECB+ is the largest and most popular dataset for cross-document Event Coreference, which is extended from ECB \cite{bejan-harabagiu-2010-unsupervised}. For each topic in ECB, \citet{cybulska2014using} add different but similar events as subtopics. We follow the same setup as previous work  \cite{cybulska2015bag, kenyon-dean-etal-2018-resolving, barhom-etal-2019-revisiting}. The  detailed statistics are shown in Table \ref{tab:ecbstat}. For both training and evaluation, we use gold event mentions. ECB+ also annotates coreference between entities that are arguments of events. We also use gold entity mentions to evaluate Entity Coreference on ECB+.

\paragraph{Preprocessing:} \enspace

\textbf{Argument generation}. ECB+ annotates arguments of each event in the same sentence, but does not annotate the role of the arguments and the event that the arguments belong to. To predict arguments for each event mention, we use AI2 SRL system ,\footnote{\url{https://demo.allennlp.org/semantic-role-labeling}} which is a reimplementation of \citet{shi2019simple}, and then we map the predicted arguments to the gold arguments. If any gold argument span overlaps with a predicted argument span, we assign the predicted role to it.

\textbf{Topic Clustering}.
Topic clustering is a common componet of cross-document coreference because it is computationally inefficient to calculate similarity of the mention pairs in all the documents. People prefer to only collect mention pairs within documents that are related. \citet{barhom-etal-2019-revisiting} implements a strong topic clustering model that uses the $K$-Means algorithm on the documents represented by TF-IDF scores of unigrams, bi-grams, and trigrams. They choose the $K$ value based on the Silhouette Coefficient method \cite{rousseeuw1987silhouettes}, and perfectly get the number of gold topics. Though there still exist wrong documents in each topic cluster, their nearly perfect clustering allows very simple baseline models to achieve very good results \cite{barhom-etal-2019-revisiting}. Since we focus on the improvement that the pairwise representation can bring, we use exactly the same topic clustering model they implemented. We use gold topics for training, and predicted topics for inference.

\paragraph{Postprocessing: Mention Clustering.} After training the pairwise coreference scorer, following previous work \cite{choubey-huang-2017-event, kenyon-dean-etal-2018-resolving, barhom-etal-2019-revisiting, catton2020stream}, we apply agglomerative clustering to the event pairs by the score from the trained scorer in Equation \ref{eq:MLP2}. Agglomerative clustering merges event clusters until no cluster pairs have a similarity score higher than a threshold. We define the cluster pair similarity score as the average score of all the event pairs across two clusters, and tune the threshold on development data. 

\paragraph{Results:} We compare with two state-of-the-art cross-document Event Coreference models using different methods: \citet{barhom-etal-2019-revisiting}, which jointly trains Entity Coreference and Event Coreference, and \citet{catton2020stream}, which jointly learns mention detection and coreference. We also compare with the same head lemma baseline implemented by \citet{barhom-etal-2019-revisiting}, which simply clusters events with same head lemma. 

To reveal the true merit of \modelname, in Table \ref{tab:ecbresult}, we separately show the effectiveness of the structured and pairwise representations as proposed in \modelname{}. In ``Unstructured'', our system only uses the trigger representation, Equation \ref{eq:finalrepentity}, to denote the representation of a pair of mention; in ``Structured'', the structured representation depicted in Equation \ref{eq:finalrep} is used; in ``Unpaired'', the representations of trigger and arguments are generated with their own sentence only instead of the concatenated two sentences; in ``Pairwise'', the representations are generated by the two concatenated sentences as described in Sec \ref{sec:model}. We see that using only structured representations improves F1 by 1.6 (from 81.3 to 82.9) from the baseline unpaired+unstructured setting, and using only pairwise representation improves F1 by 2.7 (from 81.3 to 84.0). Both 82.9 and 84.0 already outperform the state-of-the-art model \newcite{catton2020stream} on all of the evaluation metrics with large margins, particularly when using pairwise representation, 84.0 vs. 81.0 by CoNLL F1 score. When incorporating structured representation into pairwise representation, the system obtains further improvement (from 82.9 to 84.4 CoNLL F1). Please note that both \newcite{barhom-etal-2019-revisiting} and \newcite{catton2020stream} have relatively complex systems to learn event features as well as entity features. Our system only models the trigger and arguments representations given the context of two involved mentions. It clearly demonstrates the superiority of our model in learning the event-pair representation.

ECB+ also annotates coreference between entities that are arguments of events. Because entities do not have arguments, we just use \modelname~to learn the pairwise representation as Equation \ref{eq:finalrepentity}. We compare with the same two baselines: \newcite{barhom-etal-2019-revisiting} and \newcite{catton2020stream}. Both of these two baselines train their model on gold mentions, so the comparison is fair. As shown in Table \ref{tab:ecbentityresult}, our system \modelname~significantly outperforms the two baselines: 81.5 vs. 73.1.

\begin{table}[t]
\centering
\begin{tabular}{l|rrr}
\toprule
 & Train & Dev & Test \\
\midrule
Documents & 360 & 169 & 167 \\
Event mentions & 12,976 & 4,155 & 4,375 \\
Event Singletons & 5,256 & 2,709 & 2,358\\
Event Clusters & 7,460 & 3,191 & 2,963 \\

\bottomrule
\end{tabular}
\caption{KBP statistics. We use KBP2015 for \textit{train}, KBP 2016 for \textit{dev} and KBP 2017 for \textit{test}. Event Clusters include singletons.} 
\label{tab:kbpstat}

\end{table}

\begin{table*}[!t]
\centering
\begin{tabular}{lcccccccccc}
\toprule
Model & & MUC & & $\mathrm{B}^3$ & & $\mathrm{CEAF}_e$ & & BLANC && AVG-F \\ 
\midrule
\citet{huang2019improving} & \\
\enspace\enspace\enspace\enspace Predicted Mentions  && 35.66 && 43.20 && 40.02 && 32.43 && 36.75\\
\citet{lu2020e2e} & \\
\enspace\enspace\enspace\enspace Predicted Mentions  && 39.06 && 47.77 && 45.97 && 30.60 && 40.85\\
\enspace\enspace\enspace\enspace Gold Mentions && - && - && - && - && 53.72 \\
\midrule
Unpaired (Gold Mentions) && 60.23 && 52.34 && 47.44 && 45.32 && 51.33\\
\modelname~(Gold Mentions) && 63.67 && 58.41 && 54.66 && 51.72 && \textbf{57.12} \\
$\modelname_{\mathrm{BERT}}$ (Gold Mentions) && 59.11 && 53.11 && 50.6 && 45.81 && 52.16 \\


\bottomrule
\end{tabular}
\caption{Within-document event coreference performance on KBP17. Please note that the KBP15 corpus (training data) only provides trigger annotation, so we only evaluate the performance of trigger representation. ``Unpaired'' is the baseline model without pairwise representation. $\modelname_{\mathrm{BERT}}$ means this baseline model uses BERT as contextual embeddings instead of RoBERTa.} 

\label{tab:kbpresult}
\end{table*}

\subsection{Within-document Event Coreference}

Within-document event coreference focuses on event pairs in the same document, so topic clustering of documents is not needed. We use the same pairwise scorer and mention clustering algorithm described in Section \ref{sec:crossdocevent}.

We evaluate on the most widely used KBP benchmark. Similar to \newcite{huang2019improving} and \newcite{lu2020e2e}, we use the KBP 2015 dataset \cite{ellis2015overview} as training data, the KBP 2016 dataset \cite{ellis2016overview} as dev data, and the KBP 2017 \cite{Getman2015OverviewOL} as test data. The detailed statistics are shown in Table \ref{tab:kbpstat}. Because the training data KBP 2015 dataset does not have the annotation of arguments, we evaluate the performance of the representation with trigger only. 

We compare with two state-of-the-art systems on the KBP benchmark: \citet{huang2019improving}, which exploits unlabeled data to learn argument compatibility in order to improve coreference performance, and \newcite{lu2020e2e}, which jointly learns event detection and event coreference. \newcite{lu2020e2e} claims the state-of-the-art performance when predicting event coreference given predicted events, and they also report numbers using gold event mentions. Our model does not conduct mention detection, so we report our performance on gold mentions only (this is still fair since the prior SOTA system \newcite{lu2020e2e} reports on gold mentions too) and leave our numbers on predicted mentions as future work. As shown in Table \ref{tab:kbpresult}, \modelname~outperforms the unpaired baseline model with a big margin: 57.12 vs. 51.33 (on ``AVG-F''). This further verifies the effectiveness of the pairwise representation in modeling event coreference regardless of whether it is within-document or cross-document. We also need to give credit to RoBERTa that helps our simple model easily outperform the state-of-the-art model (57.12 vs. 53.72), which is a much more complicated model than ours.

\subsection{Implementation Details}
For both ECB+ and KBP models, we use $\textrm{RoBERTa}_{\textrm{Large}}$ as the encoder. The sizes of four layers of $\mathrm{MLP_1}$ are: 3076/1024/1024/1. The sizes of four layers of $\mathrm{MLP_2}$ are: 3072/1024/1024/1. We set the learning rate as 1e-06, the batch size as 32, and we run 10 epochs for training. All hyperparameters are tuned based on development data, including the threshold of agglomerative clustering.

\section{Analysis}
\label{analysis}
To further understand why pairwise representation performs much better than unpaired representation, and what limitations pairwise representation still has, we do a quantitative analysis on the errors of \modelname~and the errors of the unpaired baseline model on ECB+. For each model, we randomly sample 100 errors: 50 false negatives and 50 false positives. False negative means that the gold label of the event pair is ``coref'', but the model predicts ``not coref''. False positives mean that the gold label of the event pair is ``not coref'', but the model predicts ``coref''. We manually classify these errors into different types, and study the difference between the error distributions of the two models. 

\begin{figure*}[t]
\centering
\includegraphics[width=16cm]{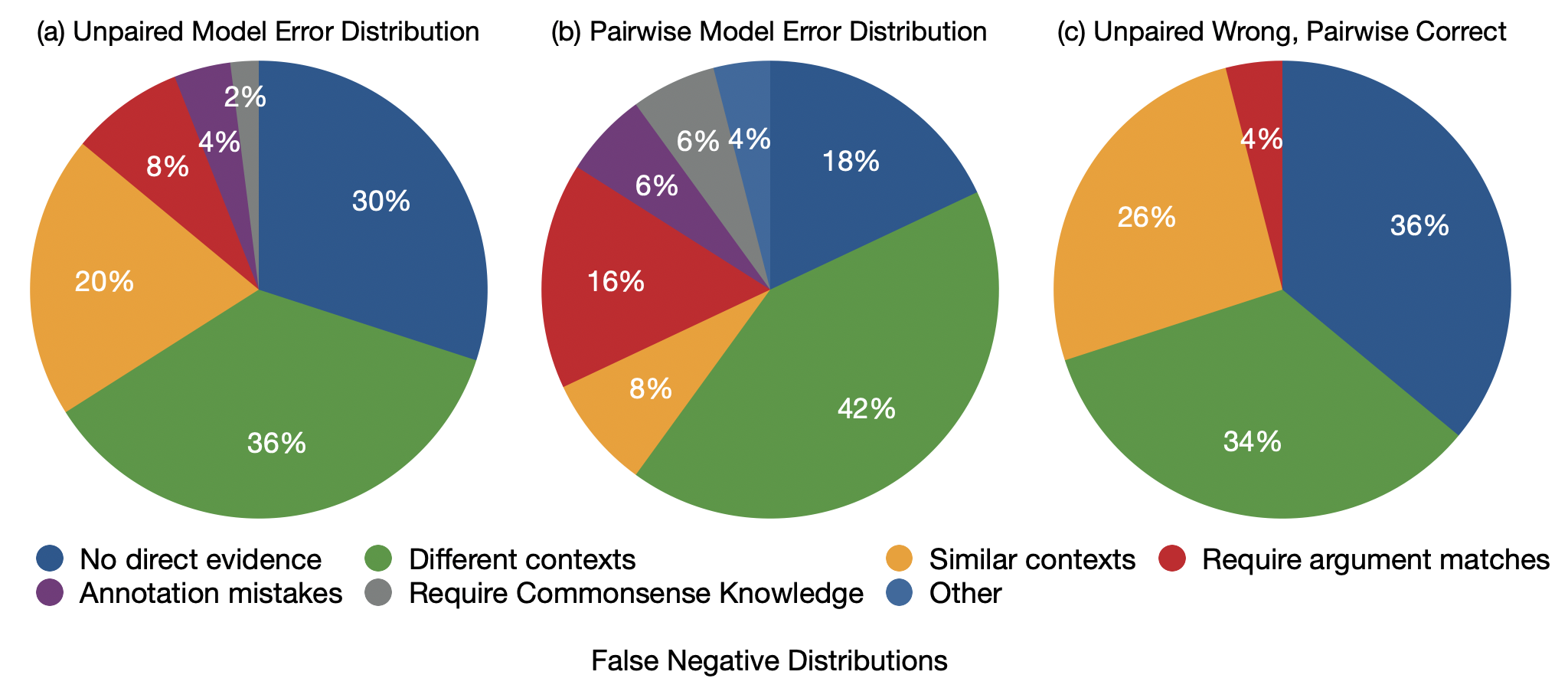}
\caption{False Negative distributions of unpaired model, and pairwise model. False negative refers to gold coreferential event pairs that the model predicts ``not coref''. More details in Sec \ref{recall}}
\label{fig:recall}
\end{figure*}

\subsection{False Negatives}
\label{recall}

Given event mention pairs with the two sentences, as listed on the bottom of Figure \ref{fig:recall}, we classify these false negatives into these 7 types: ``No direct evidence'', ``Different contexts''. ``Similar contexts'', ``Require argument matches'', ``Annotation mistakes'', ``Require commonsense knowledge'', and ``Other''.

\paragraph{``No direct evidence''} means that, just by reading the two sentences, there is no evidence in them to decide that these two mentions must be the same event. For example:\\

  \begin{minipage}{0.93\linewidth}
    {\fontsize{10}{12}\selectfont $s_1$: Smith, 26, who played a young political researcher in the show, will become the biggest star of all after \underline{\textbf{winning}} the role of the 11th Doctor.
    
    $s_2$: The guy is relatively unknown and the skeptics wondered if the right person was \underline{\textbf{chosen}}.}\\
  \end{minipage}
Just by reading these two sentences, we really do not know whether the event ``winning'' and the event ``chosen'' are same event or not. To make the correct prediction, more contexts are needed. Most prior work encoded events within only a single sentence; in this work, we use a single sentence as event context for fair comparison. As shown in Figure \ref{fig:recall}, the unpaired model has $30\%$ mistakes belong to “No direct evidence”, while the pairwise model only has $18.4\%$. This indicates that pairwise model may be more capable to learn the similarity between the context in order to make a ``guess’’ that is more likely to be correct. However, $18.4\%$ is also high. This indicates that sentence-level representation is not enough to represent an event. Event arguments usually appear in multiple sentences. Representing events in a multi-sentence level could be interesting to future work. 

\paragraph{``Different contexts''} means that the two sentences are too hard for the model to understand and there is no obvious textual similarity for the model to rely on. However, if the model understands the contexts completely, it should make the correct prediction. For example:\\

  \begin{minipage}{0.93\linewidth}
    {\fontsize{10}{12}\selectfont $s_1$: Scott Peterson has been found guilty of first-degree murder, a verdict that means he could be \underline{\textbf{executed}} if these same jurors vote as the ``conscience of their community'' that he deserves to die for his crimes.
    
    $s_2$: Laci Peterson's loved ones have ``a hole in their hearts that will never be repaired,'' a prosecutor told jurors today as he asked them to send convicted double-murderer Scott Peterson to his \underline{\textbf{death}} for killing his wife and unborn son.}\\
  \end{minipage}
 In this example, sentences are both complicated and sharing limited vocabulary, but by understanding the sentences, we can say that two event mentions are the same event. We regard this error type as hard cases, and the pairwise model suffers from these hard cases. $40.2\%$ mistakes of the pairwise model belong to hard cases ``Different contexts''. Please note that a higher ratio (40.2\% vs. 36\%) doesn't mean our pairwise model is worse than the unpaired competitor; this is because our system has resolved most of the simpler cases so the hard cases occupied the majority proportion of remaining errors. Improving the performance on complicated sentences still acts as the main challenge.
 
 \paragraph{``Similar contexts''} means that the two sentences are very similar, which should be easy for the model to make the correct prediction. For example:
 
  \begin{minipage}{0.93\linewidth}
    {\fontsize{10}{12}\selectfont $s_1$: A strong \underline{\textbf{earthquake}} struck Indonesia's Aceh province on Tuesday, killing at least one person and leaving two others missing.
    
    $s_2$: A powerful \underline{\textbf{6.1 magnitude earthquake}} hit Indonesia's Aceh province, on the island of Sumatra .}\\
  \end{minipage}
These two sentences have similar context and similar structure, which should be easy to predict two mentions as the same events. We regard this error type as easy cases. Our pairwise model reduces the error rate dramatically from 20\% to 8\% in this category, which indicates that the pairwise model is very effective to solve these simple cases. 

\paragraph{``Require argument matches''} means that to make the correct prediction, systems need to use more context or external knowledge to conduct non-trivial argument matching. For example:\\

  \begin{minipage}{0.93\linewidth}
    {\fontsize{10}{12}\selectfont $s_1$: An earthquake with a preliminary magnitude of 4.4 \underline{\textbf{struck}} in Sonoma County this morning near The Geysers, according to the U.S. Geological Survey.
    
    $s_2$: The temblor \underline{\textbf{occurred}} at 9:27 a.m. , 13 miles east of Cloverdale and 2 miles southeast of The Geysers , where geothermal forces by more than 20 power plants are harnessed to provide energy for several North Bay counties.}\\
  \end{minipage}

In order to make the correct prediction of these two sentences, the model need to realize the match between ``9:27 a.m.'' and ``this morning'', and know that ``Sonoma County'' is ``13 miles east of Cloverdale'', which requires more context or external knowledge. 

We also sample 50  errors of unpaired model where the pairwise model could predict correctly. As shown in Figure \ref{fig:recall}(c), the improvement of the pairwise representation mainly comes from better performance on ``No direct evidence'', ``Different contexts'' and ``Similar contexts''. We find that the sentences are usually very long for these errors, which suggests that the pairwise representation is better at understanding the meaning of long sentences than the unpaired representation is.

\begin{figure}[thbp]
\centering
\includegraphics[width=7.5cm]{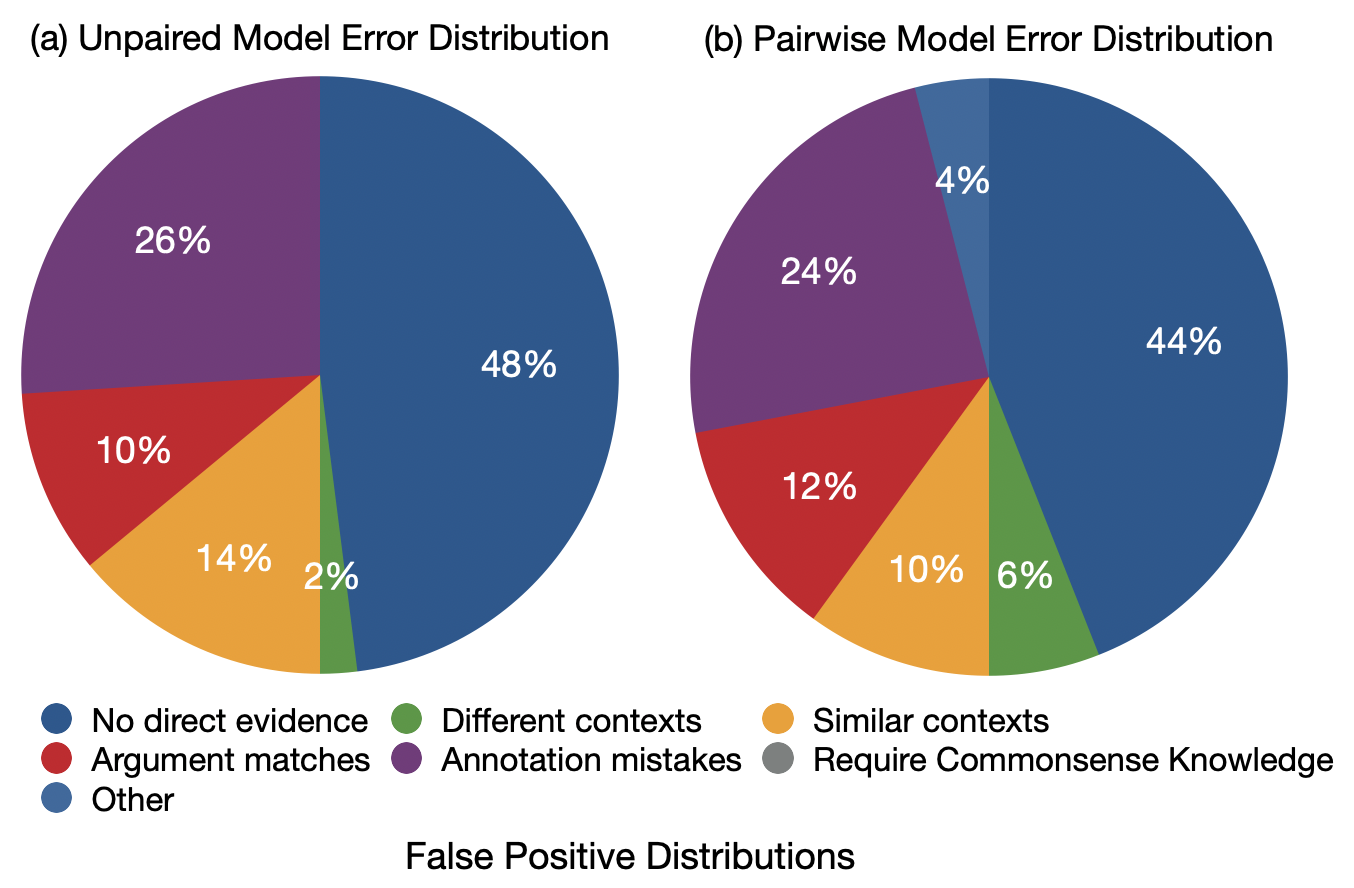}
\caption{False positive distributions of unpaired model, and pairwise model. False positive refers to gold event pairs that are not coreferential, but the model predicts ``coref''. More details in Sec \ref{precision}}
\label{fig:precision}
\end{figure}

\subsection{False Positives}
\label{precision}

For the sampled false positives, we also manually classify them into 7 types same as the types of false negatives. The only difference is that, now ``Similar contexts'' become hard cases, and ``Different contexts'' become easy cases. As shown in Figure \ref{fig:precision}, for both the unpaired model and the pairwise model, most of the precision errors belong to ``No direct evidence'' and ``Annotation mistakes''. After carefully studying these errors, we find that it is actually very hard to determine that two mentions are not the same event. For example:\\

  \begin{minipage}{0.93\linewidth}
    {\fontsize{10}{12}\selectfont $s_1$: Four bombs were dropped within just a few moments - two \underline{\textbf{landed}} inside the camp itself, while the other two bombs were dropped near the airstrip where a UN helicopter was delivering much needed food aid.
    
    $s_2$: "Two of the bombs \underline{\textbf{fell}} within the Yida camp , including one close to the school," said UNHCR spokesman Adrian Edwards .}\\
  \end{minipage}
By understanding these two sentences, we think, without knowing whether ``the camp itself'' in the first sentence is the same camp as ``Yida camp'' in the second sentence, it is impossible to make the correct prediction. The gold label for this pair is ``not coref'', so we can only classify it to ``No direct evidence''. We think that these errors again emphasize that the representation of events should be multi-sentences level instead of sentence level. We only use SRL to find event arguments, which limits arguments to be in the same sentences. We think that it may be essential to find events across sentences in future works.

We also find that there exist some errors that we think are \textit{annotation mistakes}. For example:\\
    
  \begin{minipage}{0.93\linewidth}
    {\fontsize{10}{12}\selectfont $s_1$: Smith, 26, who played a young political researcher in the show, will become the biggest star of all after \underline{\textbf{winning}} the role of the 11th Doctor .
    
    $s_2$: The BBC says little-known actor Matt Smith will \underline{\textbf{take over}} the title role in the long-running sci-fi series ``Doctor Who.''}\\
  \end{minipage}
We do not see any reasons that these two mentions are not the same event, but if there are other contexts indicating that they are not the same event, this error would be classified to ``No direct evidence''. So in conclusion, to further improve the performance on false positives, longer-range context will be needed.

\section{Conclusion}
In this work, we propose a simple 
representation learning scheme, \modelname, for event coreference. \modelname~learns a  mention-pair representation by forwarding concatenated sentences into RoBERTa, where sentences provide the context of mentions. This representation is applied to both within-document and cross-document event coreference benchmarks and obtains state-of-the-art performance. In addition, we augment this pairwise representation with structured argument features to further improve its performance. 

\section*{Acknowledgments}

This work was supported by Contract FA8750-19-2-1004 with the US Defense Advanced Research Projects Agency (DARPA), the Oﬃce of the Director of National Intelligence (ODNI), Intelligence Advanced Research Projects Activity (IARPA), via IARPA Contract No. 2019-19051600006 under the BETTER Program, and a Focused Award from Google. Approved for Public Release, Distribution Unlimited. The views and conclusions contained herein are those of the authors and should not be interpreted as necessarily representing the oﬃcial policies, either expressed or implied, of ODNI, IARPA, the Department of Defense, or the U.S. Government. The U.S. Government is authorized to reproduce and distribute reprints for governmental purposes notwithstanding any copyright annotation therein.

\bibliographystyle{acl_natbib}
\bibliography{anthology,cited,new,custom}

\appendix

\end{document}